\documentclass[conference]{IEEEtran}
\IEEEoverridecommandlockouts
\usepackage{cite}
\usepackage{amsmath,amssymb,amsfonts}
\usepackage{algorithmic}
\usepackage{graphicx}
\usepackage{textcomp}
\usepackage{xcolor}
\usepackage{multirow}
\def\BibTeX{{\rm B\kern-.05em{\sc i\kern-.025em b}\kern-.08em
    T\kern-.1667em\lower.7ex\hbox{E}\kern-.125emX}}

\usepackage{algorithm}
\usepackage{algorithmic}
\usepackage{multirow}
\usepackage{tablefootnote}
\usepackage{caption}
\usepackage{subcaption}

\newcommand{\ours}{\text{MA3C}}
\newcommand{\softmax}{\text{softmax}}

\begin{document}

\title{Multi-agent Attention Actor-Critic Algorithm for  Load Balancing in Cellular Networks}

\author{
    \IEEEauthorblockN{Jikun Kang, Di Wu, Ju Wang, Ekram Hossain, Xue Liu, Gregory Dedek}
    \IEEEauthorblockA{Samsung Electronics, Canada
    \\ \{jikun.kang, e.hossain\}@partner.samsung.com, \{di.wu1, j.wang1,  steve.liu, greg.dudek\}@samsung.com}
}

\maketitle

\begin{abstract}
In cellular networks, User Equipment (UE)  handoff from one Base Station (BS) to another, giving rise to the load balancing problem among the BSs.  To address this problem, BSs can work collaboratively to deliver a smooth migration (or handoff) and satisfy the UEs' service requirements.
This paper formulates the load balancing problem as a Markov game and proposes a Robust \textbf{M}ulti-agent \textbf{A}ttention \textbf{A}ctor-\textbf{C}ritic (Robust-MA3C) algorithm that can facilitate collaboration among the BSs (i.e., agents). In particular, to solve the Markov game and find a Nash equilibrium policy, we embrace the idea  of adopting a nature agent to model the system uncertainty.
Moreover, we utilize the self-attention mechanism, which encourages high-performance BSs to assist low-performance BSs.
In addition, we consider two types of schemes, which can facilitate load balancing for both active UEs and idle UEs.
We carry out extensive evaluations by simulations, and simulation results illustrate that, compared to the state-of-the-art MARL methods, Robust-\ours~scheme~can improve the overall performance by up to 45\%.

\begin{IEEEkeywords}
Load balancing, multi-agent reinforcement learning, Nash equilibrium, self-attention
\end{IEEEkeywords}

\end{abstract}

\section{Introduction}

In a mobile cellular system,  User Equipment (UE) movements across Base Stations (BSs) result in a  Load Balancing (LB) problem causing UE dissatisfaction \cite{DBLP:conf/vtc/Chiaraviglio0BF20}.
To balance the load, UEs are migrated among BSs and channels in each BS.
Thus, it is crucial to coordinate multiple base stations to reallocate UEs when they move from one region to another.
In a 5G network,  BSs could be more close to each other compared with LTE/4G network, which means UEs are more likely to move from one BS to another more frequently.
This makes BSs' collaboration even more important than before.

Recently, Reinforcement learning (RL) approaches have shown the advantages of solving the load balancing problem \cite{amal2022}.
However, existing RL-based LB algorithms cannot address multi-BSs LB problem properly.
The single agent solution is proposed to control BSs with one policy \cite{DBLP:journals/corr/abs-2111-00008}.
Hierarchical \cite{DBLP:conf/icc/KangCWXLDLP21} and transfer learning-based RL methods \cite{DBLP:conf/globecom/WuKXLLCRJLPLD21} have also been developed and showed improved performance in terms of data throughput and load variation reduction.
However, these methods fail to consider multi-agent interactions and cannot guarantee Nash equilibrium.
In addition, the classical multi-agent actor-critic algorithm \cite{DBLP:conf/globecom/MaiYXGN20} does not consider Nash equilibrium.

\begin{figure}[t]
    \centering
    \includegraphics[width=0.8\linewidth]{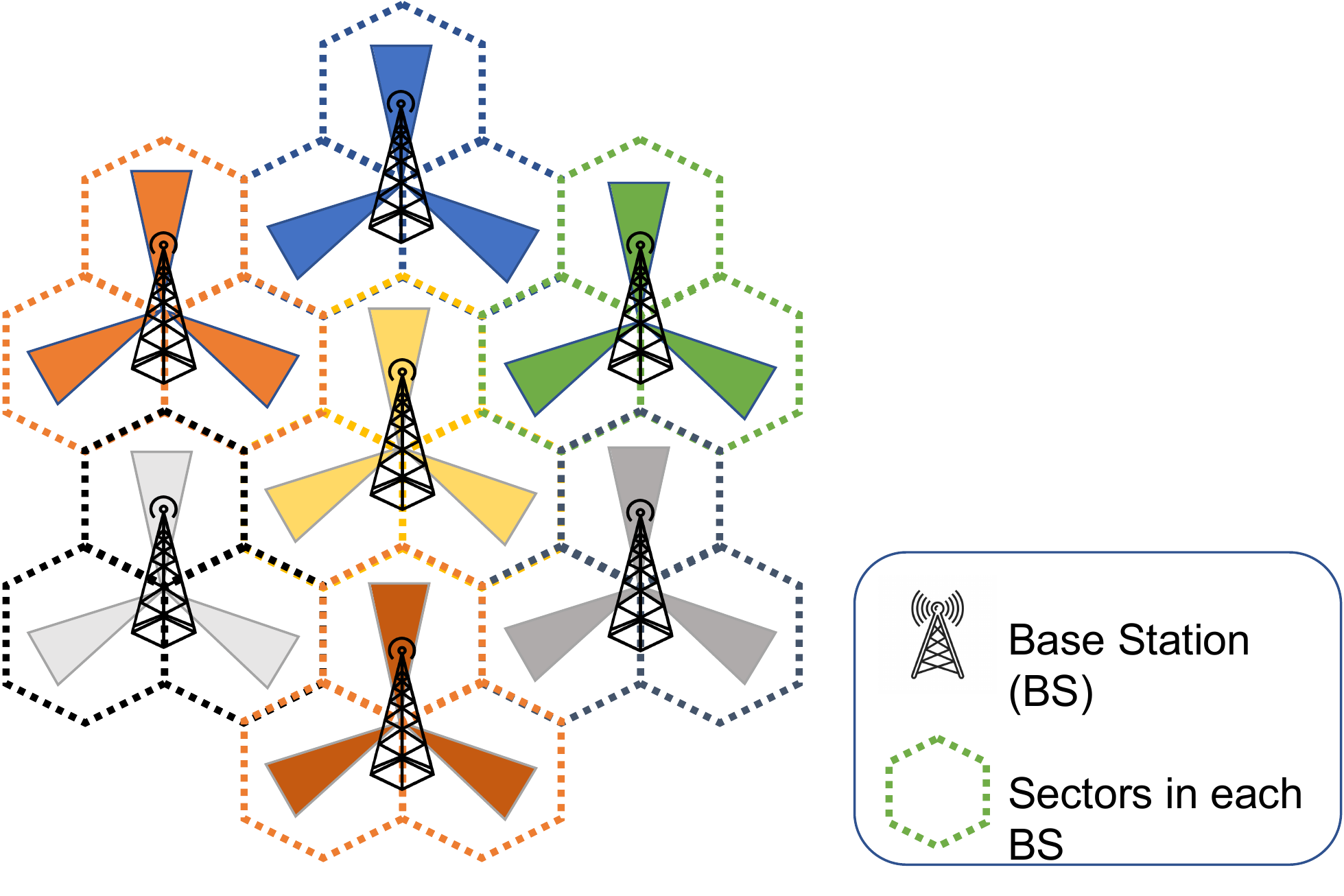}
    \caption{A cellular network tessellation with 7 BSs. Each hexagon denotes one sector. One BS consist of three sectors, which draw in the same color. Each sector controls 120 angle degrees. Furthermore, in each sector, there are 4 channels residing on 4 different carrier frequencies. In this paper, we control the parameters of these 7 BSs  to achieve load balancing.}
    \label{fig:simulator}
\end{figure}

To address aforementioned issues, in this paper, we formulate the load balancing problem over multiple base stations (Fig.~\ref{fig:simulator}) as a Markov game.
Specifically, we treat base stations as agents and learn the corresponding policies to control the  load balancing.
To solve the Markov game, we propose Robust-\ours, a Robust-\textbf{M}ulti-agent \textbf{A}ttention \textbf{A}ctor-\textbf{C}ritic algorithm.
The main focus of our work is to leverage collaboration among BSs to improve system performance.
In particular, we seek to enhance the UE experience in underperforming BSs.
A major challenge in achieving these goals is that agents tend to maximize their own goals greedily, which affect the overall experiences of the UEs.
The state-of-the-art rule-based method cannot address this challenge properly, since they do not formulate the problem in multi-agent settings.
The single-agent approach available in the literature is not suitable to foster BS collaboration~\cite{DBLP:journals/corr/abs-2111-00008}.
This is because, in single-agent algorithms, BSs are controlled by one centralized neural network, which fails to converge when the action space increases with the number of agents.
The state-of-the-art MARL load balancing algorithms are not able to deal with this challenge either, since they spare no effort to preserve the UE experience in poorly performing BSs.

To overcome this challenge, we adopt the following two mechanisms.
First, to encourage collaboration among all BSs, we introduce an attention-based \cite{DBLP:conf/nips/VaswaniSPUJGKP17} multi-agent actor critic algorithm.
Specifically, during the centralized training, each transmitted information will be assigned a weight value.
The motivation is that agents will have different attention to the received information.
Since our objective is to maximize the overall system performance, high-performance agents can help the poorly performing agents based on transmitted information.
Second, we utilize the nature agent idea \cite{DBLP:conf/nips/ZhangSTGMB20} to achieve the Nash equilibrium policy.
The nature agent plays against all other agents by selecting the worst-case actions.
Thus, to fight against the worst-case scenario raised by the nature agent, all agents need to work together and develop a joint equilibrium policy.

We evaluate the proposed Robust-\ours~method against the state-of-the-art (SOTA) multi-agent RL (MARL)-based methods in a system-level network simulator for a network, which is depicted in Fig.~\ref{fig:simulator}.
The simulation results show that our method can improve the network throughput by up to 45\%.

In summary, this paper makes the following contributions:
\begin{itemize}
    \item We formulate the multi-BS LB problem as a Markov game and propose a MARL solution, which enhances system performance by exploiting collaborations among the BSs. 
    \item To improve the underperformaing BSs' performance, we propose an attention-based weighting mechanism to each transmitted information.
    \item To model uncertainty, we add a nature agent to play against all agents to find the Nash equilibrium policy.
    \item We benchmark the proposed method against five other methods and demonstrate its performance superiority. 
\end{itemize}

\section{Related Work}


Load balancing algorithms have been designed for wireless networks to evenly distribute UEs across different network resources, i.e., frequency bands (channels) and base stations.
The MAS model \cite{DBLP:conf/cse/KimHLY09} is for a dynamic load balancing scheme for a multi-agent system in which the agent selection is based on agents' credit value, and location selection is based on the inter-machine communication load.
Another mobility-based load balancing infrastructure is called JIAC \cite{DBLP:conf/seke/StenderKA06}. The migration decision is taken locally which prevents the centralized control problems. 
In \cite{myint2005framework}, a centralized load balancing system is given. The selection policy is based on the job’s execution time, while the location policy is based on negotiation with cluster nodes. 
Previous work \cite{DBLP:journals/corr/abs-2111-00008} has explored single-agent RL for load balancing in network systems, with a close-to-heuristic performance achieved in moderate-scale simulations.
This paper applies and studies MARL algorithms on the network load balancing problem in a realistic testbed, which considers load balancing for idle UEs and active UEs at the same time.

Approaches that address multi-agent cooperative tasks can be generally grouped into two types. 
The first type is based on implicit message exchange.
These types of methods utilize a centralized training framework to share agents' value function parameters, which facilitates agents' collaboration.
MADDPG \cite{DBLP:conf/nips/LoweWTHAM17} adopts global observations and actions for each action-value function estimation.
COMA \cite{DBLP:conf/aaai/FoersterFANW18} leverages the advantage of this idea and estimates each advantage function using a counterfactual baseline.
The second type of methods is based on explicit communication using learned protocols \cite{DBLP:conf/nips/FoersterAFW16, DBLP:conf/iclr/ChuCK20, DBLP:conf/nips/SukhbaatarSF16}. 
In DIAL \cite{DBLP:conf/nips/FoersterAFW16}, each agent generates message and Q-value function at the same time.
Each agent then encoded the learned message and computed it with other transmitted information.
CommNet \cite{DBLP:conf/nips/SukhbaatarSF16} learns a generalized communication protocol, which aggregates all of the received messages and computes the average value.
NeurComm \cite{DBLP:conf/iclr/ChuCK20} propose a differentiable communication protocol, which can reduce information loss.
In this paper, we focus on the first type with implicit message exchange, because it is more applicable for a real-world scenario without additional communication protocol requirements.

\section{System Model an Assumptions}

\subsection{Wireless Network Terminologies}

We use the term ``load" to refer to the number of UEs being served.
The term ``Base Station" (BS) describes a physical site, where radio access devices are placed.
In each BS, there are three non-overlapping ``sectors", each of which controls 120 angle degrees.
A sector serves the UEs located in a certain direction which is associated with the corresponding BS.
Each sector contains several ``channels" corresponding to the carrier frequencies supported by a sector.
A channel is a service entity that serves UEs in a certain direction and on a specific carrier frequency.

\subsection{Communication Load Balancing Features}
In this work, we focus on leveraging reinforcement learning to optimize load balancing for two types of users, i.e., idle UEs and active UEs. The idle UE load balancing (IULB) is a load balancing feature designed specifically for idle UEs.  This load balancing technique can help adjust the host channels for idle UEs from a crowded channel to a less overloaded channel.  The channel is referred to as a combination of the sector and frequency. The control knobs for IULB are the channel re-selection ratios, which lie between 0 and 1.  By adjusting the re-selection ratios, the probabilities of frequency to be selected as the host channel will be adjusted. Then, when the status for UEs changes from idle to active, the communication load will be distributed more evenly between different cells. 
Active UE load balancing (AULB) is a load balancing feature designed for active UEs that are actively exchanging data with the BS. The AULB will be triggered when the load of the current serving channel is larger than the neighbouring channel, and one of the neighbouring channels will be chosen. For AULB, there will be two threshold values for each channel, which help to determine when the load balancing feature will be determined and which neighbouring channel will be determined as the target channel. 

\subsection{System Metrics}
\label{subsec:metrics}

Suppose that there are $N_U$ UEs (either active, idle, or mobile) in the network.
Define $U_i$ as the set of UEs associated with the $i$-th channel.
Among $U_i$, there are both active UEs (denoted as $U_i^a$) and idle UEs (denoted as $U_i^d$).
Naturally, we have $U_i = U_i^a \cup U_i^d$.
Further, let $u_{i,k}$ denote the $k$-th UE in the $i$-th channel.
Note that an idle UE at the current moment may become active in the future, and vice versa.
We aim to balance the assignments of UEs to different channels, to improve the system performance in terms of the following performance metrics.

\textbf{Average Throughput $G_{aver}= \frac{1}{N_U}\sum_{i}\sum_{k} \frac{A_{i,k}}{T}$} measures the overall system performance, where $T$ is the time period of interest, and $A_{i,k}$ denotes the total size of packets received by $u_{i,k}$ within $T$.
Improving this metric means increasing the overall system performance.

\textbf{Minimum Throughput $G_{min}= \min_{i,k}\left(\frac{A_{i,k}}{T}\right)$} captures the worst-case UE performance.

\textbf{Standard Deviation} $G_{sd} = \left(\sqrt{\frac{1}{N_U}\sum_{i}\sum_{k} (\frac{A_{i,k}}{T} - G_{aver})^2}\right)^{-1}$ represents the fairness services to all UEs. 
Minimizing this metric reduces the gap between different UEs' performance, and improves fairness to UEs.

We linearly combine these metrics as the reward $r=G_{aver}+G_{min}-G_{sd}$.
Note that these metrics are in the same range, which does not require normalization.

\section{Methodology}

\subsection{Definitions and Problem Statement}

There exist $N$ BSs, which allocate UEs to different channels based on their current location and usage.
All BSs work together to satisfy the UEs' communication requirements.
In this work, we formulate this multi-base-station load balancing task as a Markov game $G$, which is comprised of a set of states and possible actions. 
Each action transits the current state to a new state and receives a reward. The objective is to find the optimal policy, i.e., an action in each state to maximize the expected total reward. 
A Markov game can be described by a tuple $G=(\mathcal{N}, \mathcal{S}, \mathcal{A}, \mathcal{P}, \mathcal{R})$:
\begin{itemize}
\item {\em $\mathcal{N}$}: the set of agents in the Markov game. Here, we denote them as $\mathcal{N} = \{1,\ldots,N\}$, where $N$ is the total number of base stations.

\item {\em $\mathcal{S}$}: the set of joint states. We denote agent $k$'s state as $s^k$. 
In our case, there are three types of states, which are number of UEs in every BS $\boldsymbol{s_{ue}}$, the bandwidth utilization of every BS $\boldsymbol{s_{band}}$, and the average throughput of every BS $\boldsymbol{s_{tput}}$.

\item {\em $\mathcal{A}$}: the set of joint actions of all agents, $a = \{a^1 \times\cdots\times a^N\}$. Agent $k$'s action is denoted by $a^i$. In particular, every action $a^k$ contains two components. The first component regulates the active UE load balancing operations and the second component controls the idle UE load balancing parameters.

\item {\em $\mathcal{P}$}: the state transition probability function, where $P(S_{t+1} = s'|S_t=s,A_t=a)$ maps a state--action pair at time $t$ to a probability distribution over states at time $t+1$, such that $P: S\times A \times S' \rightarrow [0,1]$. 

\item {\em $\mathcal{R}$}: the reward function. In the multi-agent cooperation setting, the goal is to maximize the total reward $r = r^1 + \ldots + r^n$, where $R^k: S^k \times A^k \rightarrow R^k$ is the reward of agent $i$ (to be defined in Sec.~\ref{subsec:metrics}).

\end{itemize}

Besides, we denote by $\pi=\{\pi^1,\ldots,\pi^N\}$ the set of policies of all agents.
Formally, we define the multi-agent LB problem as follows:
\begin{equation}\label{eq:problem_def}
    \mathcal{J(\pi)}=\max_{\boldsymbol{\alpha^k},\boldsymbol{\beta^k},\boldsymbol{\gamma^k}\sim \pi^k} \sum_{k=1}^N G_k,
    \vspace{-3mm}
\end{equation}
\begin{align}
    \text{s.t.  } & \boldsymbol{\alpha^k} \in [\boldsymbol{\alpha_{min}^k}, \boldsymbol{\alpha_{max}^k}],\\
    & \boldsymbol{\beta^k} \in [\boldsymbol{\beta_{min}^k}, \boldsymbol{\beta_{max}^k}],\\
    & \boldsymbol{\gamma^k} \in [\boldsymbol{\gamma_{min}^k}, \boldsymbol{\gamma_{max}^k}],
\end{align}
where $G_k$ denotes the system performance of BS $k$, $\boldsymbol{\alpha_{min}^k}$ and $\boldsymbol{\alpha_{max}^k}$ denote the controllable range of AULB actions of BS $k$, and $\boldsymbol{\beta_{min}^k}$, $\boldsymbol{\beta_{max}^k}$, $\boldsymbol{\gamma_{min}^k}$ and $\boldsymbol{\gamma_{max}^k}$ define the controllable range of IULB actions of BS $k$. 
In this paper, $\boldsymbol{\alpha^k}\in[-2 db, 2 db], \boldsymbol{\beta^k}\in[-20 db, 20 db], \boldsymbol{\gamma^k}\in[-20 db, 20 db], k=\{1,\ldots,N\}$.
{\em Due to its combinatorial nature, solving this multi-agent LB problem using traditional optimization methods will not be feasible for a practical system.} 

The goal of MARL is to learn a set of agent policies $\{[\boldsymbol{\alpha_{i,j}^k},\boldsymbol{\beta_{i,j}^k},\boldsymbol{\gamma_{i,j}^k}] \sim \pi^k\}_{k=1,\ldots,N}$ that maximize the total expected return $\sum_{k=1}^N G_k$ per episode.

For multi-agent load balancing, since we focus on maximizing the joint reward of all BSs, the proposed solution should be in accordance with the Nash equilibrium.
In particular, the Nash equilibrium is defined as follows:
\begin{equation}
\label{eqn:nash}
    \mathcal{J}(\pi^k_*,\pi^{-k}_*)\geq \mathcal{J}(\pi^k_*,\pi^{-k}_*), \forall k\in \mathcal{N},
\end{equation}
where $\pi^{-k}= \prod_{k\neq i}\pi^i$ refers to other agents' joint policies except agent $k$.
Our goal is to find a Nash equilibrium such that, given all other agents' equilibrium policies $\pi^{-k}_*$, there is no motivation for agent $k$ to deviate from $\pi^k_*$.

\subsection{Actor-Critic Algorithm}
To start with, we first introduce the actor-critic algorithm, which is the primitive of the state-of-the-art algorithms in MARL \cite{DBLP:conf/nips/LoweWTHAM17, DBLP:conf/aaai/FoersterFANW18}.
The actor-critic algorithm \cite{DBLP:conf/nips/KondaT99}, as its name suggests, consists of two functions: the actor function and the critic function.
The critic function estimates the action value, which is often noted as the Q-value function.
This function serves as the policy learning guidance as it estimates the performance of the current action, which can be seen as an expert that controls the direction of gradient learning.
The actor function directly controls the agent's behaviors according to the ``suggestion'' from a critic.

\subsection{Multi-agent Actor-Critic Algorithm}

We propose to extend the actor-critic algorithm to the multi-agent scenario.
We adopt the centralized critic training decentralized actor execution framework \cite{DBLP:conf/nips/LoweWTHAM17, DBLP:conf/aaai/FoersterFANW18}.
During the critic training phase, each agent receives neighbor agents' information (the size of the transmitted information is about a few bytes) at every step $t$, which are states $\boldsymbol{s}_i$ and actions $\boldsymbol{a}_i$.
Thus, the critic loss function can be written as:
\begin{equation}
    \begin{aligned}
\mathcal{L}(\phi^i) &= \mathbb{E}_{s,a,r}[y-Q^i(\boldsymbol{s^1_t},\boldsymbol{a^1_t},\ldots, \boldsymbol{s^k_t},\boldsymbol{a^k_t};\phi^i)]^2,\\
\end{aligned}
\end{equation}
\label{eqn:origin_critic_loss}
$\text{where~} y = r^i + \gamma Q^i(\boldsymbol{s^1_{t+1}},\boldsymbol{a^1_{t+1}},\ldots, \boldsymbol{s^k_{t+1}},\boldsymbol{a^k_{t+1}};\phi^i)$.
While in the execution phase, the agent makes decisions based on its states.
According to the policy gradient theorem \cite{DBLP:conf/nips/SuttonMSM99}, the actor updating function can be written as:
\[
    \label{eqn:origin_actor_loss}
    \nabla \mathcal{J}(\theta^i) = \mathbb{E}_{s^i,a^i\sim \mathcal{D}}[\nabla_{\theta^i} \text{log}\pi^i(\boldsymbol{a^i}|\boldsymbol{s^i})Q^i(\boldsymbol{s^1_t},\boldsymbol{a^1_t},\ldots, \boldsymbol{s^k_t},\boldsymbol{a^k_t})],
\]

\subsection{Attention-Based Multi-agent Actor-Critic Algorithm}
To encourage collaboration among different agents, we propose an attention-based multi-agent learning approach.
Specifically, in the centralized critic training phase, we adopt the self-attention mechanism \cite{DBLP:conf/nips/VaswaniSPUJGKP17} to learn the importance weights of transmitted information.
For the simplicity of calculation, we encode information $\boldsymbol{e}^i = g(\boldsymbol{s}^i, \boldsymbol{a}^i)$, where $g(\cdot)$ is a multi-layer perceptron (MLP).
Hence, the importance weight of agent $j$ to agent $i$ is denoted as $\alpha^{i,j} = \softmax\left(\frac{\boldsymbol{e^i}\cdot \boldsymbol{e^j}}{\sqrt{d_k}}\right)\boldsymbol{e^i}$,
where $d_k$ is the dimension of $\boldsymbol{e^k}$.
Then, we assign an attention weight to each received information $e^j$ and re-write the critic loss function as follows:
\begin{equation}
\label{eqn:critic_loss}
\begin{aligned}
\mathcal{L}(\phi^i) &= \frac{1}{N}\sum_j(y-Q^i(\boldsymbol{s^i}, \alpha^{i,1}\boldsymbol{e^1},\ldots,\alpha^{i,N}\boldsymbol{e^j};\phi^i))^2,\\
\text{where~} y &= r^i + \gamma Q^i(\boldsymbol{e^i}, \alpha^{i,1}\boldsymbol{e^1},\ldots\alpha^{i,N}\boldsymbol{e^N};\phi^i),
\end{aligned}
\end{equation}
where $Q^i(\cdot)$ is the centralized action-value function \cite{DBLP:conf/nips/LoweWTHAM17}, to which we extend the neighbor attention. 
The inputs of this function are encoded by agents' observations and actions. 
The outputs are action-value for agent $i$. 

Moreover, with the introduced attention weights, the actor updating function can be expressed as:
\begin{equation}
    \label{eqn:actor_loss}
    \begin{aligned}
    \nabla \mathcal{L}(\theta^i) = \mathbb{E}_{s^i,a^i\sim \mathcal{D}}[&\nabla_{\theta^i} \text{log}\pi^i(\boldsymbol{a^i}|\boldsymbol{s^i})\\&Q^i(\boldsymbol{e^i}, \alpha^{i,1}\boldsymbol{e^1},\ldots\alpha^{i,N}\boldsymbol{e^N})].
    \end{aligned}
\end{equation}



\subsection{Improving Nash Equilibrium by Adding a Nature agent}

\begin{figure}[t]
    \centering
    \includegraphics[width=0.9\linewidth]{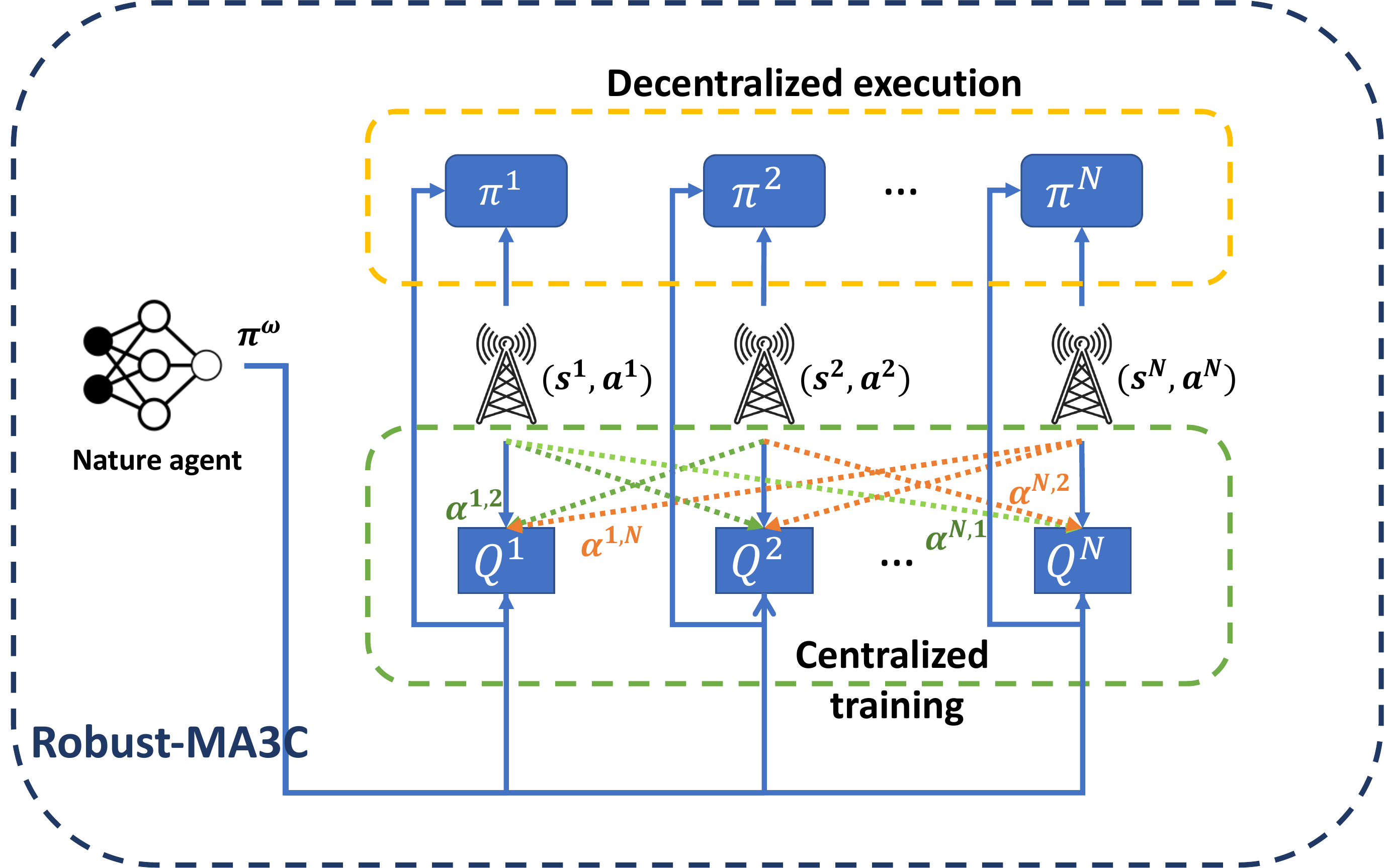}
    \caption{Illustration of proposed algorithms. Dash lines refer to the information transmission. Different dash line colors correspond to different importance weights.}
    \label{fig:archi}
\end{figure}

To further enhance agents' collaborations and solve the Nash equilibrium defined in Eqn.~\ref{eqn:nash}, we adopt the nature agent idea \cite{DBLP:conf/nips/ZhangSTGMB20} and propose a corresponding Robust-\ours~algorithm.
The motivation is that the nature agent always plays against each agent by selecting the worst-case actions at every state, which can be also seen as the model of the reward uncertainty.
Thus, to fight against the worst-case scenario raised by the nature agent, all agents need to work together and develop a joint equilibrium policy.

We replace the reward with the nature agent in the critic function and develop a robust-critic loss function:
\begin{equation}
\label{eqn:robust-critic_loss}
\begin{aligned}
y = \pi^{\omega}(\hat{\boldsymbol{s}},\boldsymbol{\hat{a}}) + \gamma Q^i(\boldsymbol{e^i}, \alpha^{i,1}\boldsymbol{e^1},\ldots\alpha^{i,N}\boldsymbol{e^N};\phi^i)
\end{aligned},
\end{equation}
where nature-agent's policy is approximated by an neural network parameterized by $\omega$, $\hat{\boldsymbol{s}}$ stands for the nature-agent's state, and $\boldsymbol{\hat{a}}$ refers to nature-agent's action.
We update the nature agents policy parameter as follows:
\begin{equation}
\label{eqn:update_omega}
    \omega_{t+1} = \omega_t-\beta_t\cdot\nabla\pi_{\omega_t}(\hat{\boldsymbol{s_t}},\boldsymbol{\hat{a}_t}),
\end{equation}
where $\beta_t$ is the learning rate that diminishes over time, i.e., $\lim_{t\rightarrow \infty}\beta_t=0$. 
The Robust-\ours~algorithm is summarized in \textbf{Algorithm ~\ref{alg:robust-ours}}.
An overview of the proposed method, i.e., Robust-\ours~is given in Fig.~\ref{fig:archi}.

\begin{algorithm}[hbtp]
\caption{Robust-\ours}
\label{alg:robust-ours}
\begin{algorithmic}[1]
\FOR{$t=1$ to maximum episode length}
\STATE Execute actions $a=(a_1,\ldots,a_N)$ and observe reward $r$ and next states $s'=(s'_1,\ldots,s'_N)$.
\STATE Store $(s,a,r,s')$ in replay buffer $\mathcal{D}$
\FOR{agent $i$ to $N$}
\STATE Sample a random minibatch of samples (s,a,r,s') from $\mathcal{D}$
\STATE Update the nature-agent's policies by using Eqn~\ref{eqn:update_omega}.
\STATE Update critic loss by using Eqn.~\ref{eqn:robust-critic_loss}.
\STATE Update actor loss by using Eqn.~\ref{eqn:actor_loss}. 
\ENDFOR
\ENDFOR
\end{algorithmic}
\end{algorithm}

\section{Evaluation}
\label{sec:eval}

\label{subsec:eval_sectors}
\begin{figure*}[hbtp]
     \centering
     \begin{subfigure}[b]{0.32\textwidth}
         \centering
         \includegraphics[width=\textwidth]{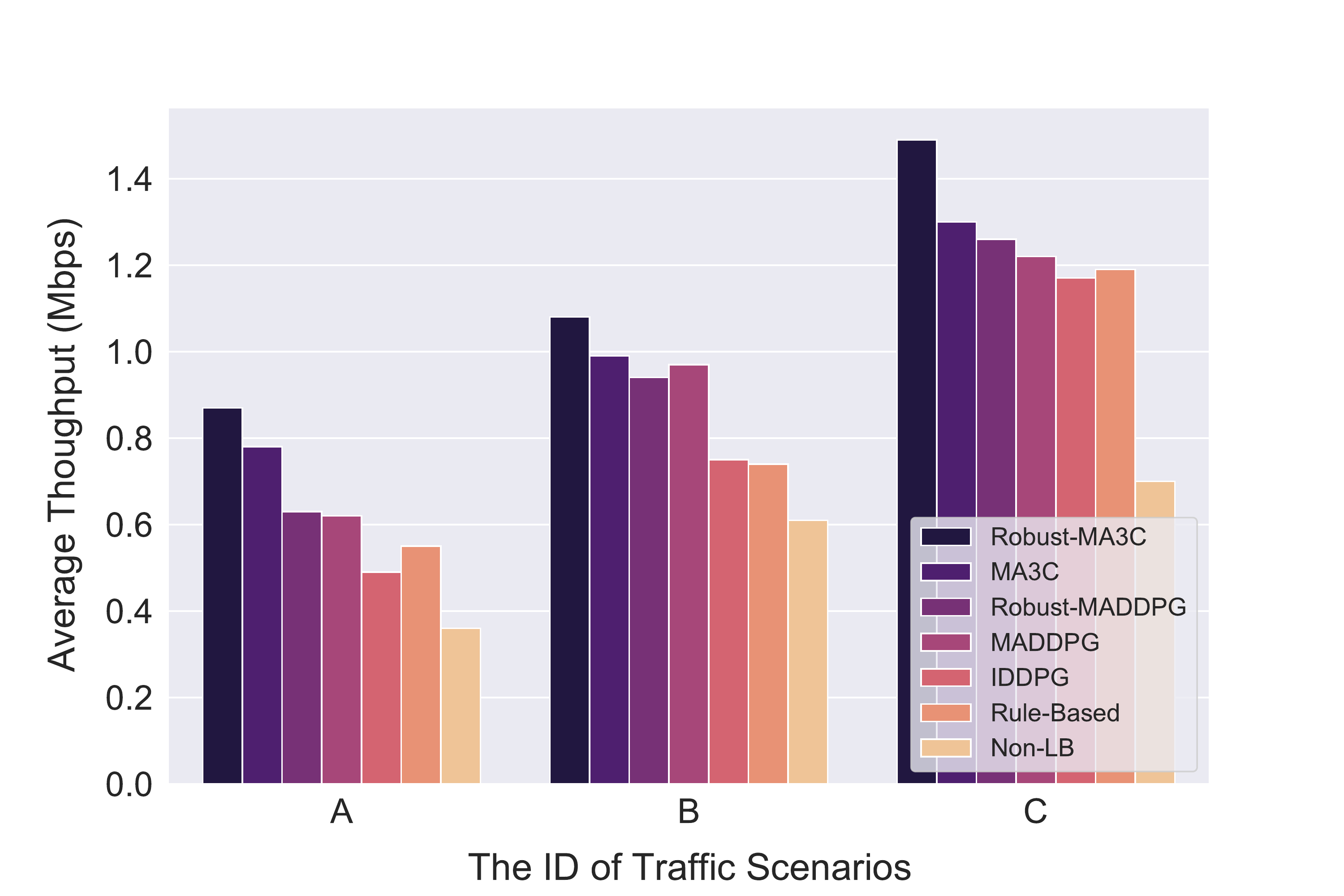}
         \caption{Average Throughput}
         \label{fig:aver_sec}
     \end{subfigure}
     \hfill
     \begin{subfigure}[b]{0.32\textwidth}
         \centering
         \includegraphics[width=\textwidth]{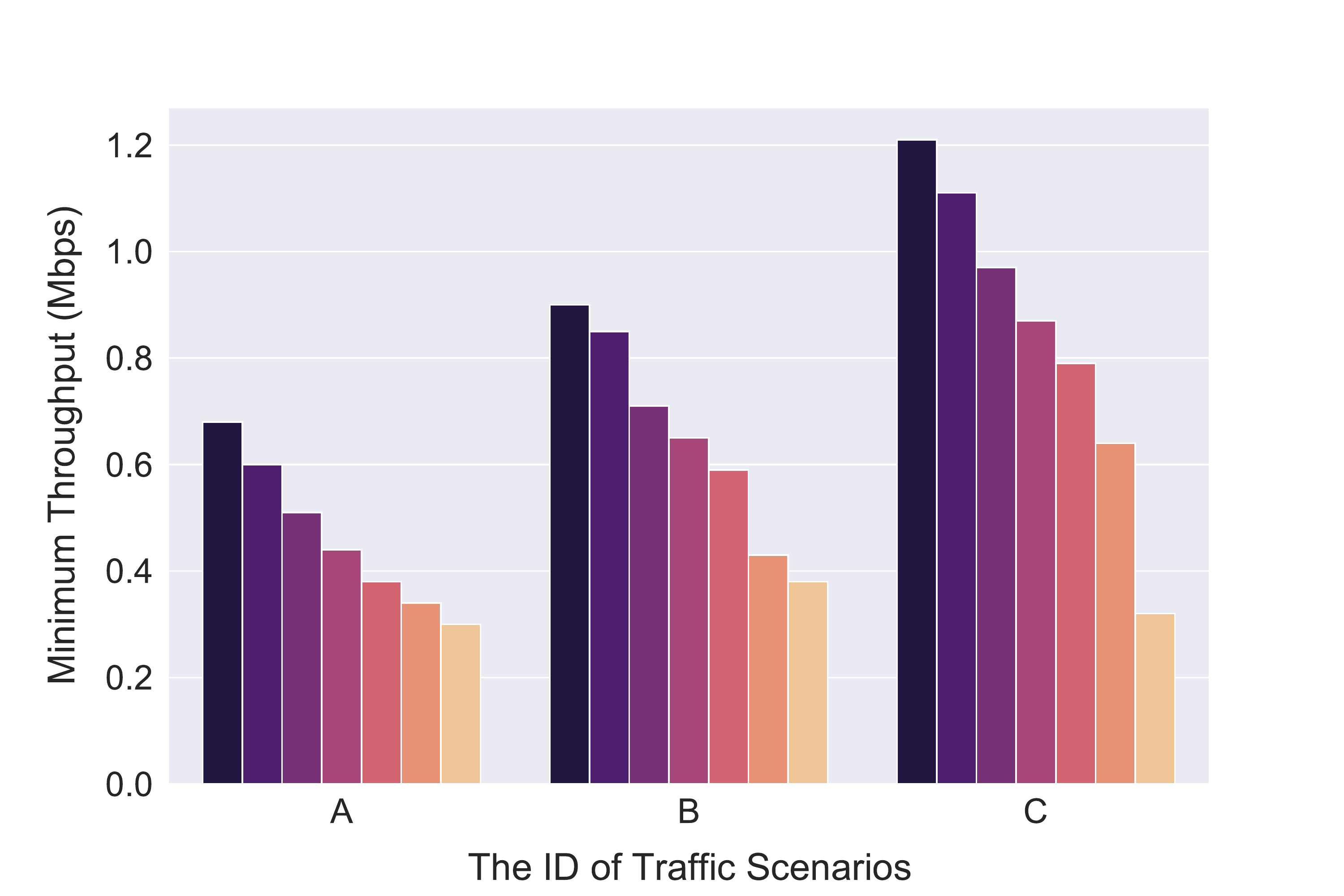}
         \caption{Minimum Throughput}
         \label{fig:min_sec}
     \end{subfigure}
     \hfill
     \begin{subfigure}[b]{0.32\textwidth}
         \centering
         \includegraphics[width=\textwidth]{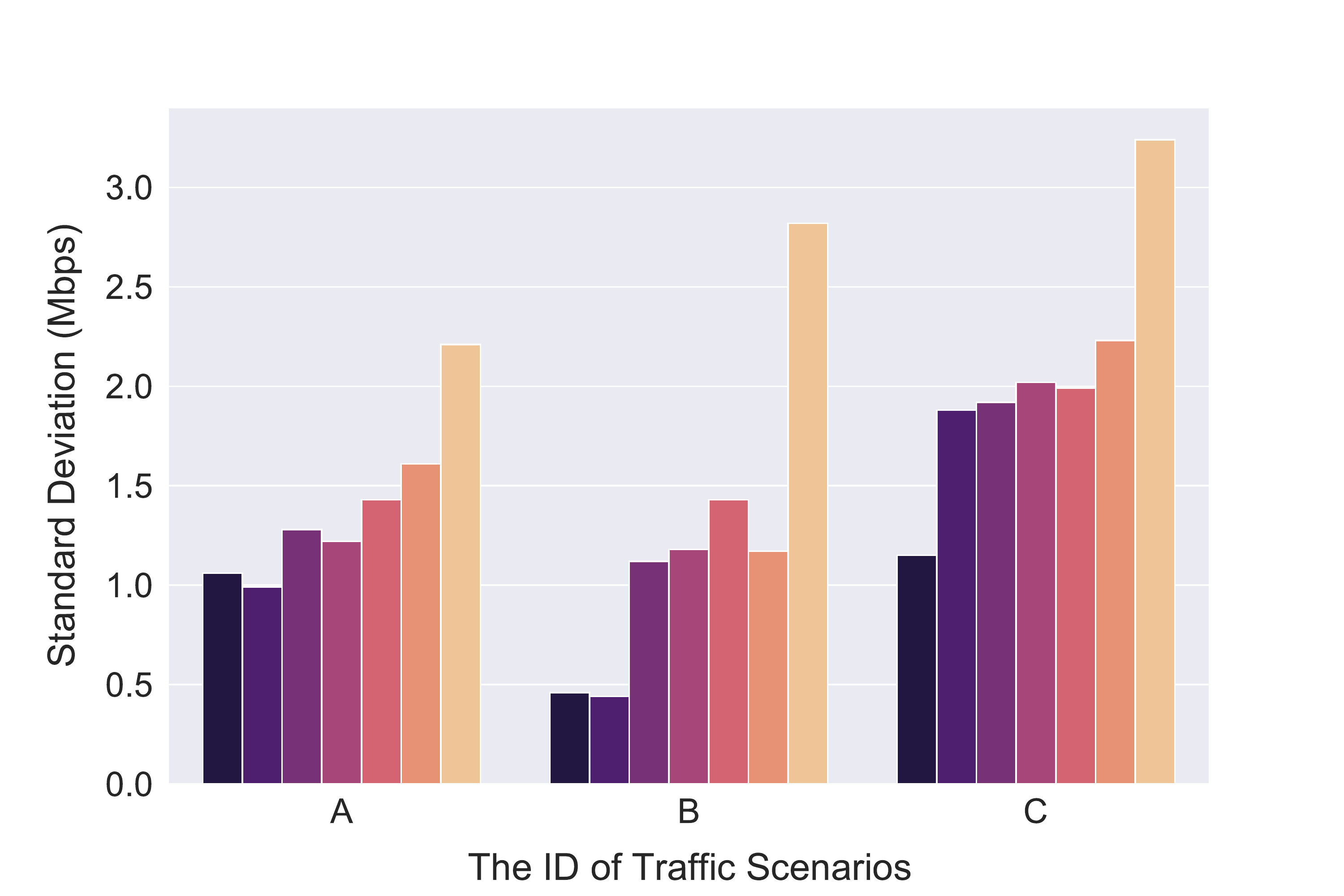}
         \caption{Standard Deviation}
         \label{fig:sd_sec}
     \end{subfigure}
        \caption{System performance over various metrics in different traffic scenarios. 
        In metric average throughput, the higher the better. In minimum throughput, the higher the better. In standard deviation, the lower the better.}
        \label{fig:sectors}
\end{figure*}

\subsection{Experiment Setup}

The experiments reported here utilize a proprietary system-level network simulator.
This simulator was created to simulate the behaviors of cellular communication networks. 
We use a network with 7 BSs to benchmark the proposed method, each of which supports 3 sectors.
In each sector, there are 4 channels residing on 4 different carrier frequencies.
(These 4 carrier frequencies are identical across different sectors and BSs.)
The scenario is wrapped around at the edges.
We emulate various traffic scenarios and days.
The details of these simulation parameters can be seen in Table.~\ref{tb:params}.
Specifically, two types of UEs need to be balanced, with the two aforementioned load balancing mechanisms, i.e., AULB and IULB.
These UEs are uniformly distributed geographically at initialization.
The active UE movement follows a random walk process with an average speed of $3m/s$.
The packet arrival follows a Poisson process with an average inter-arrival time of $200ms$.
\begin{table}[htbp]
\centering
\begin{tabular}{c|ccccc}
\hline
Traffic ID         & Day & UE & Active UE & Idle UE & Packet Size \\ \hline
A                  & 1   & 40 & 12        & 28      & 0.41 Mbps   \\ \hline
B                  & 1   & 40 & 10        & 30      & 1.03 Mbps   \\ \hline
\multirow{3}{*}{C} & 1   & 40 & 7         & 33      & 0.89 Mbps   \\ \cline{2-6} 
                   & 2   & 40 & 18        & 22      & 1.18 Mbps   \\ \cline{2-6} 
                   & 3   & 40 & 19        & 21      & 1.11 Mbps   \\ \hline
\end{tabular}
\caption{Parameters of different traffic scenarios and days.}
\label{tb:params}
\end{table}

To demonstrate the effectiveness of the proposed method, we conduct experiments in a scenario illustrated in Fig~\ref{fig:simulator}.
In this scenario, there are 7 BSs, 6 arranged in a hexagonal ring around a central station.
In this scenario, actions taken by one BS affect other BSs' decisions.
We control all of these 7 BSs to analyze the effectiveness of the proposed method.

\subsection{Methods Evaluated}
We implement the following baseline methods. 
\textbf{Non-LB} evaluates the system performance without load balancing.
\textbf{Rule-based} controls a BS based on prior knowledge and fixed control parameters.
\textbf{Independent DDPG} trains each BS independently by utilizing DDPG policy learning algorithm \cite{DBLP:journals/corr/LillicrapHPHETS15}.
\textbf{MADDPG} introduces centralized training decentralized execution framework to facilitate agents' collaboration, which is one of the SOTA MARL methods \cite{DBLP:conf/nips/LoweWTHAM17}.
\textbf{Robust-MADDPG} develops a Q-learning algorithm to find robust Nash equilibrium policies, which also solves the model uncertainty problem \cite{DBLP:conf/nips/ZhangSTGMB20}. This is a follow-up work of MADDPG.
\textbf{\ours~(ours)} the ablation study of proposed Robust-MA3C without a robust agent. Note that this approach does not guarantee to converge to the Nash equilibrium.

\subsection{Evaluation Results in Different Traffic Scenarios}

We evaluate all the comparison schemes on different traffic scenarios, which are shown in Fig.~\ref{fig:sectors} and showcase the evaluation results on the aforementioned metrics. 
From this figure, we can observe that the proposed \ours~and Robust-\ours~algorithm consistently outperforms other baselines in terms of various metrics. 
Compared to the third best baseline (i.e., Robust-MADDPG), \ours~increases the performance of average throughput, minimum throughput, and standard deviation by up to 12\%, 31\%, and 42\%.
This illustrates that, by utilizing the neighbor-aware attention mechanism, our proposed \ours ~MARL algorithm enhances agents' collaborations.

Furthermore, we visualize the effectiveness of combining robust agents by comparing Robust-\ours  ~to the third best baseline.
The results show that Robust-\ours ~improves the performance over corresponding metrics by up to 18\%, 37\%, and 45\%.
This verifies the effectiveness of finding Nash equilibrium and further advances the state-of-the-art.

\subsection{Evaluation Results in Different Days}

\begin{figure*}[hbtp]
     \centering
     \begin{subfigure}[b]{0.32\textwidth}
         \centering
         \includegraphics[width=\textwidth]{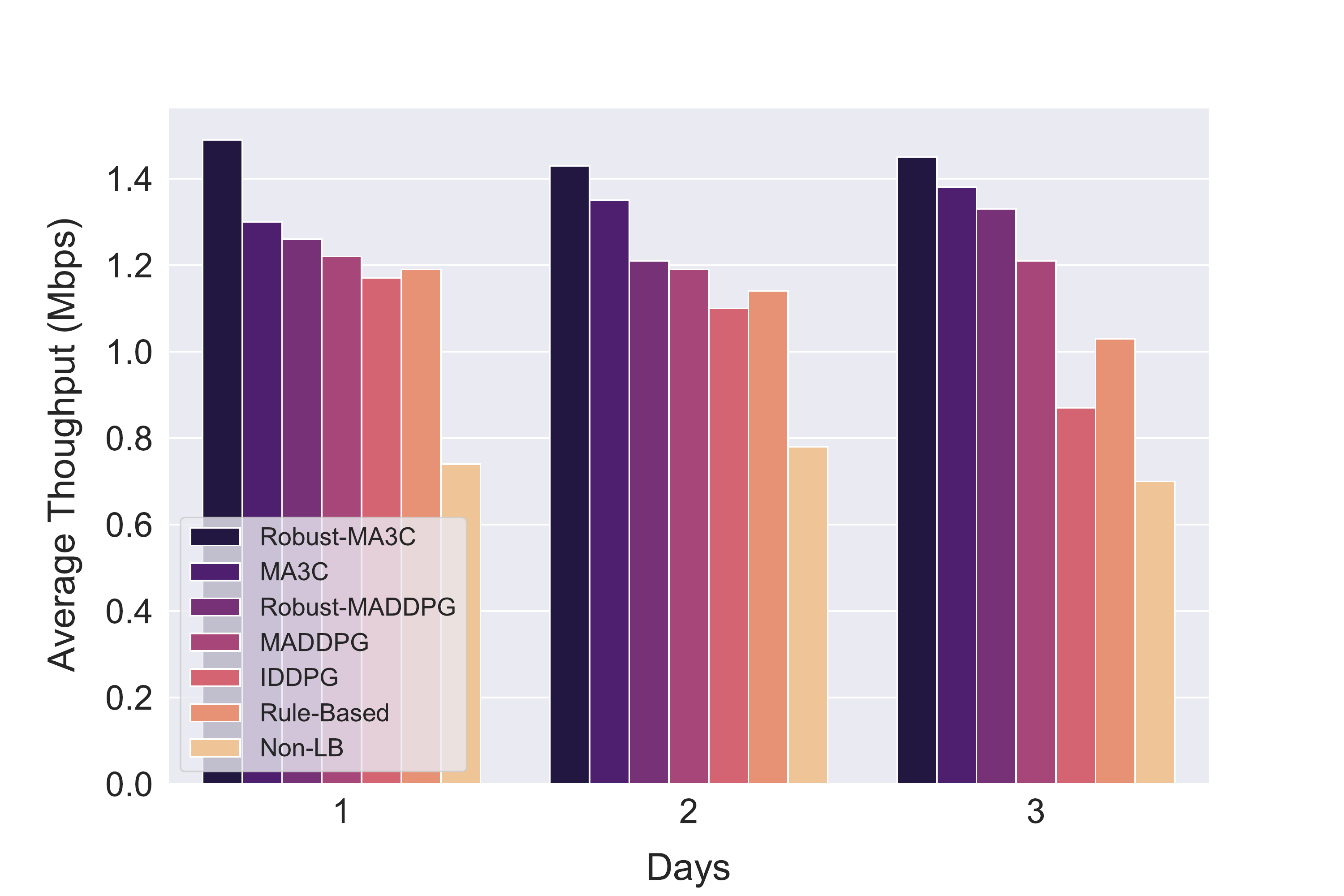}
         \caption{Average Throughput}
         \label{fig:aver}
     \end{subfigure}
     \hfill
     \begin{subfigure}[b]{0.32\textwidth}
         \centering
         \includegraphics[width=\textwidth]{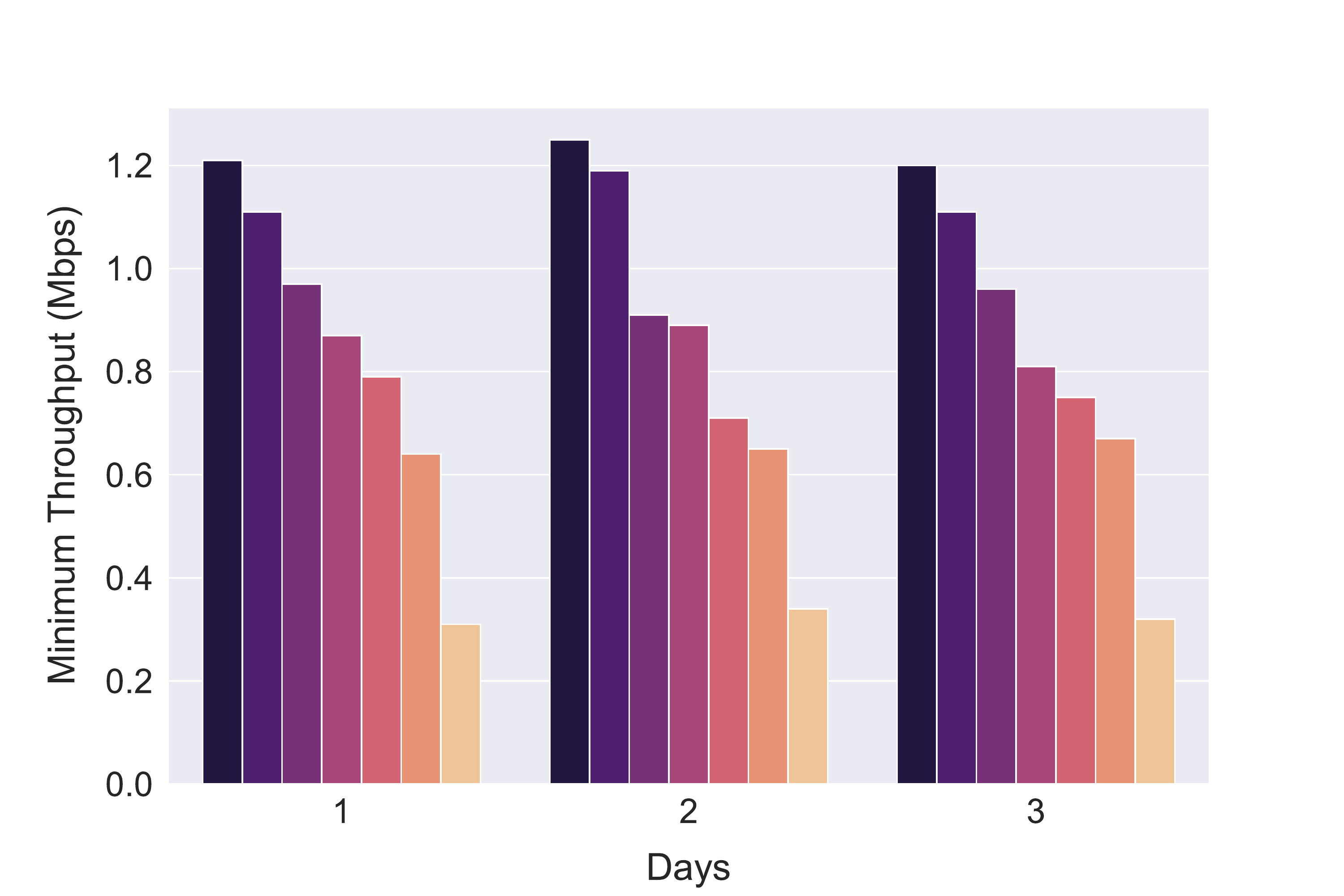}
         \caption{Minimum Throughput}
         \label{fig:min}
     \end{subfigure}
     \hfill
     \begin{subfigure}[b]{0.32\textwidth}
         \centering
         \includegraphics[width=\textwidth]{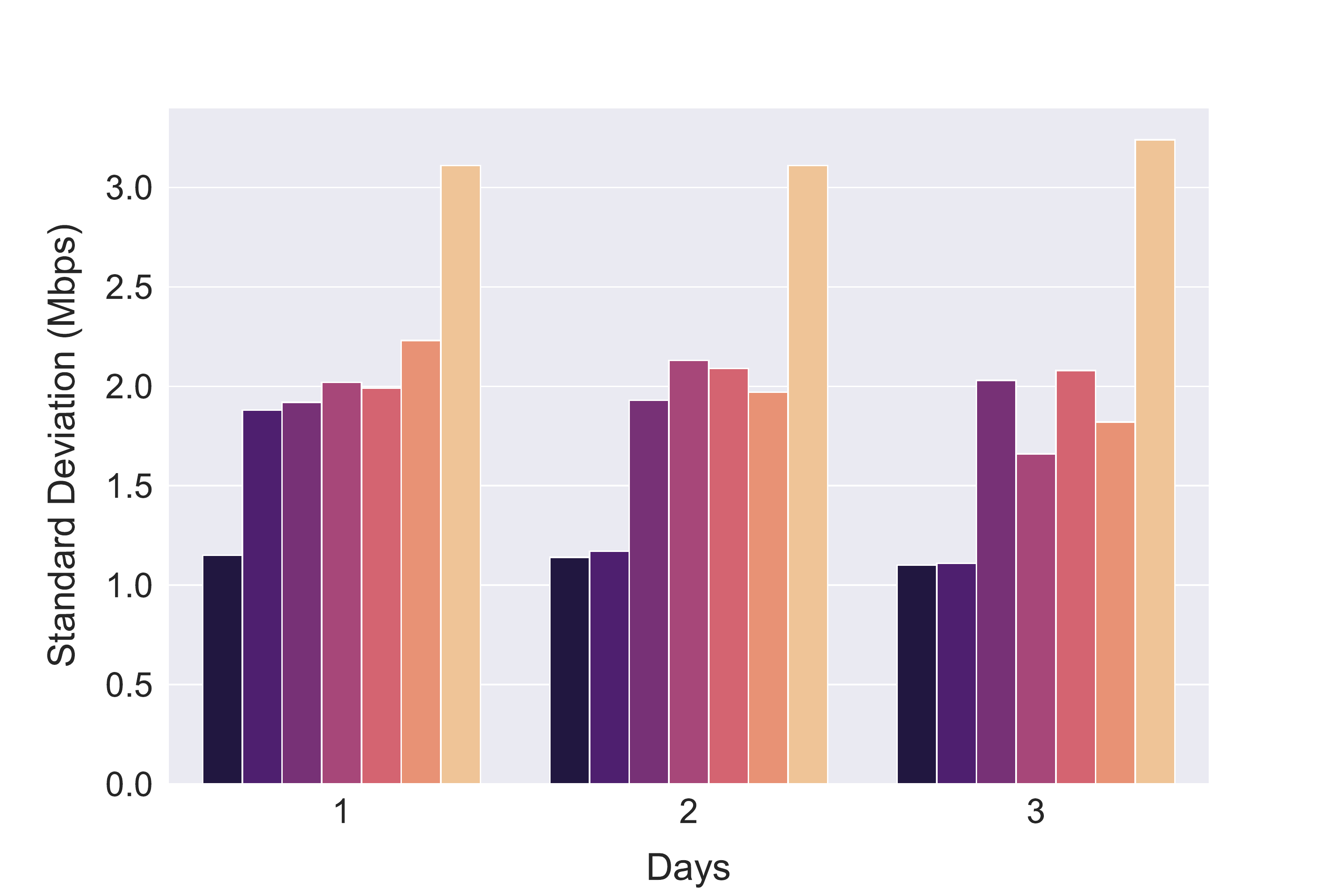}
         \caption{Standard Deviation}
         \label{fig:sd}
     \end{subfigure}
        \caption{System performance in different days.}
        \label{fig:days}
\end{figure*}

Next, we further present results on different days of one traffic scenario, which are depicted in Fig.~\ref{fig:days}.
We draw the same conclusion that the proposed two algorithms can improve the system performance overall metrics.
Precisely, \ours~increases the performance of average throughput, minimum throughput, and standard deviation by up to 12\%, 31\%, and 42\%.
As for Robust-\ours, it improves the performance over metrics by up to 19\%, 37\%, and 45\%.
These results further illustrate the effectiveness of the proposed algorithms. 

\subsection{Evaluations of Individual BSs}

\begin{figure}[hbtp]
    \centering
    \includegraphics[width=\linewidth]{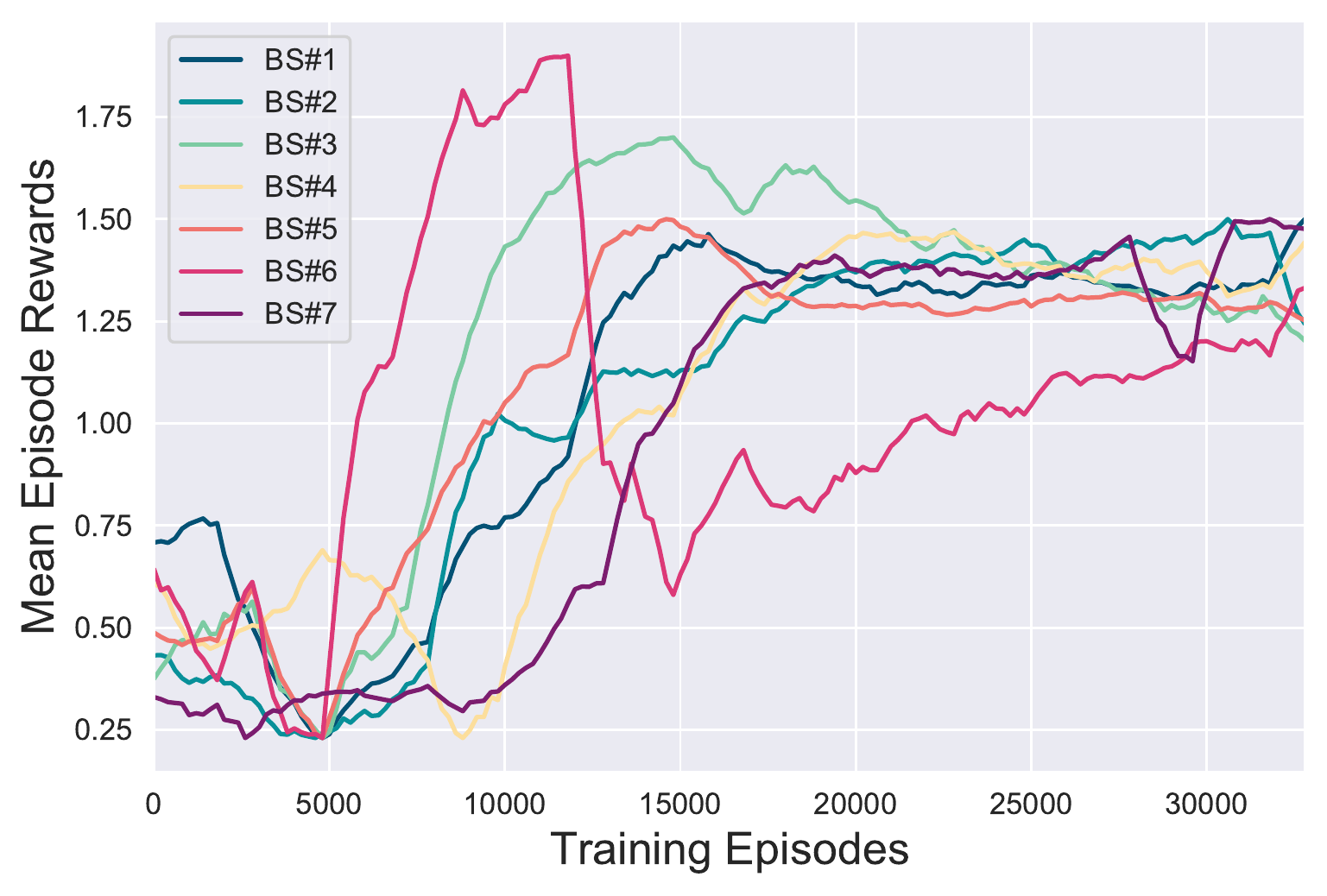}
    \caption{The \ours~learning curves of seven BSs. The x-axis refers to the training episodes and the y-axis stands for the mean episode rewards that defined in Sec.~\ref{subsec:metrics}.}
    \label{fig:robust-ma3c_agents}
\end{figure}


To better understand the performance of each BS, we present a learning curve in Fig.~\ref{fig:robust-ma3c_agents}.
Note that we use the Robust-\ours~as the training algorithm as an example since \ours ~illustrates the same phenomenon.
In this figure, we observe that all BSs converge to the same range of rewards $1.25\sim 1.5$.
This result illustrates that after training around 30k episodes, all BSs tend to achieve a Nash equilibrium and further illustrates the effectiveness of the proposed method.
Among the learning curves, we can observe \textbf{an interesting phenomenon:} to reach a Nash equilibrium, some agents that achieve high reward at the early stage would decrease their performance for poorly performing BSs.
For example, we find there is a big drop in the learning curve of BS\#6. This
BS\#6 achieves a high reward at around 12k episodes.
Then, the performance of this agent drastically drops to a low value.
This happens because other agents cannot find optimal policies at this time.
To help the poorly performing agents, BS\#6 would like to explore some sub-optimal actions and sacrifice its performance to help the other agents. 
Finally, the received reward of BS\#6 steadily increases after other agents converge.
With the help of good agents, all agents collaborate and converge on the same range of rewards.

\section{Conclusion}

We have studied the multi-agent load balancing problem in cellular networks, where active UEs and idle UEs are migrated across BSs.
A major challenge lies in how to enable collaboration among multiple BSs to achieve the joint Nash equilibrium policy for the agents.
To address this challenge, we have proposed a novel algorithm, i.e., Robust-\ours, a Robust-\textbf{M}ulti-agent \textbf{A}ttention \textbf{A}ctor-\textbf{C}ritic load balancing algorithm.
Specifically, to provide a good migrating service, we utilize an attention mechanism to improve the performance of low-performance agents with the help of high-performance agents.
In addition, we leverage the idea of nature-agent to achieve the Nash equilibrium.
The evaluation results demonstrate that Robust-\ours can achieve superior performance over the rule-based and other multi-agent reinforcement learning algorithms on three key network performance metrics.
By using the simulation results, we have also demonstrated how the agents converge to Nash equilibrium.

\bibliographystyle{ieeetr}
\bibliography{ref.bib}

\end{document}